\documentclass[]{spie}  

\usepackage{amsmath,amsfonts,amssymb}
\usepackage{graphicx}
\usepackage[colorlinks=true, allcolors=blue]{hyperref}

\graphicspath{ {./images/} }
\usepackage{setspace}
\usepackage{tocloft}
\usepackage{mathtools}
\usepackage{gensymb}
\usepackage[font=small,labelfont=bf,labelsep=space]{caption}
\title{Validation of object detection in UAV-based images using synthetic data \footnote{ This a pre-publication draft of a paper that is published in the Proceedings of SPIE Defense and Commercial Sensing. The final version of the paper is available from the SPIE digital library. Please cite as: E.-J. Lee, D. Conover, H. Kwon, S. S. Bhattacharyya, J. Hill, and K. Evensen, "Validation of object detection in UAV-based images using synthetic data", Proceedings Volume 11746, Artificial Intelligence and Machine Learning for Multi-Domain Operations Applications III; 117462A (2021) https://doi.org/10.1117/12.2586860.}}

\author[a]{Eung-Joo Lee}
\author[b]{Damon M. Conover}
\author[a]{Shuvra S. Bhattacharyya}
\author[b]{Heesung Kwon} 
\author[c]{\\Jason Hill}
\author[c]{Kenneth Evensen}
\affil[a]{University of Maryland, ECE Department and UMIACS, College Park, MD 20742, USA}
\affil[b]{DEVCOM Army Research Laboratory, Adelphi, MD}
\affil[c]{Defense Threat Reduction Agency (DTRA), Fort Belvoir, VA}

\authorinfo{Further author information: (Send correspondence to Eung-Joo Lee)\\Eung-Joo Lee.: E-mail: eungjoo.y.lee@gmail.com}

\begin{document} 
\maketitle

\begin{abstract}
Object detection is increasingly used onboard Unmanned Aerial Vehicles (UAV) for various applications; however, the machine learning (ML) models for UAV-based detection are often validated using data curated for tasks unrelated to the UAV application. This is a concern because training neural networks on large-scale benchmarks have shown excellent capability in generic object detection tasks, yet conventional training approaches can lead to large inference errors for UAV-based images. Such errors arise due to differences in imaging conditions between images from UAVs and images in training. To overcome this problem, we characterize boundary conditions of ML models, beyond which the models exhibit rapid degradation in detection accuracy. Our work is focused on understanding the impact of different UAV-based imaging conditions on detection performance by using synthetic data generated using a game engine. Properties of the game engine are exploited to populate the synthetic datasets with realistic and annotated images. Specifically, it enables the fine control of various parameters, such as camera position, view angle, illumination conditions, and object pose. Using the synthetic datasets, we analyze detection accuracy in different imaging conditions as a function of the above parameters. We use three well-known neural network models with different model complexity in our work. In our experiment, we observe and quantify the following: 1) how detection accuracy drops as the camera moves toward the nadir-view region; 2) how detection accuracy varies depending on different object poses, and 3) the degree to which the robustness of the models changes as illumination conditions vary.
\end{abstract}

\keywords{Object detection, UAV-based image, Synthetic data generation, Domain adaptation, Model validation.}

\begin{spacing}{2}   

\section{Introduction}
\label{sect:intro}  
Rapid and accurate detection of objects of interest from onboard Unmanned Aerial Vehicles (UAV) has become an increasingly important component of practical solutions for various applications \cite{semsch2009autonomous, honkavaara2013processing, erdelj2016uav, vidalmata2019bridging}. Although machine learning models recently have shown successful results in generic object detection tasks due to large-scale benchmarks  \cite{farhadi2018yolov3, lin2017focal, lee2019me, ren2015faster}, detection in UAV-based images has unique challenges \cite{wu2019delving}. This is primarily because of significant differences in imaging conditions between images from UAVs \cite{du2018unmanned, zhu2018visdrone, bozcan2020air, barekatain2017okutama} and images in the large-scale benchmarks used for training \cite{everingham2015pascal, lin2014microsoft, deng2009imagenet}. To obtain accurate detection results, the training dataset needs to be large and diverse and represent the scenes in which the UAVs will fly \cite{narayanan2018real}.

For example, a training set should contain thousands of images captured at different distances, viewing angles, orientations, and under different illumination conditions with various targets and backgrounds. To rigorously account for all combinations of these attributes at high levels of granularity is a laborious and costly endeavor. Additionally, manually annotating each image requires additional costs.

To tackle this problem, we focus on an alternative approach that can effectively substitute for the large-scale collection of real data. As an alternative to large-scale UAV-based data collections, synthetic data can be generated using a game engine \cite{tremblay2018training, kar2019meta}. The game engine that we used is the Unity Real-Time Development Platform, which enables the creation of scenes using various terrains and people 3D models \cite{saleh2018effective, yan2020x1}. Within Unity, we also have the flexibility to adjust the visibility and lighting within the scene. In this study, we demonstrate that a synthetic dataset can be used to validate machine learning models.

\begin{figure}[!htbp]
\centering
\includegraphics[width=0.26\textwidth]{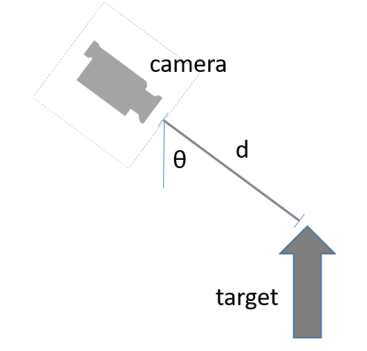}
\caption{The relationship between the camera and target object}
\label{fig:camera}
\end{figure}

\begin{figure}[!htbp]
\centering
\includegraphics[width=0.30\textwidth]{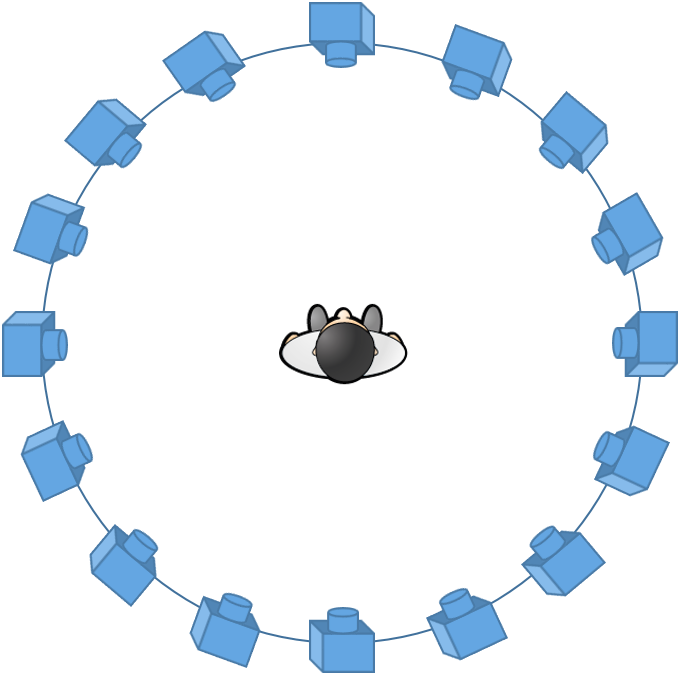}
\caption{Full 360° collections of a stationary target by moving a single camera}
\label{fig:circle}
\end{figure}

Using our generated synthetic dataset, we characterize the performance of multiple machine learning models, with different levels of architectural and model complexity, with respect to the camera position. This allows us to visualize the expected performance for each classifier for a given distance ($\it{d}$) and pitch angle ($\theta$) between the camera and the target (see Figure ~\ref{fig:camera}), as the camera circles and remains pointed inward at the target, as shown in Figure ~\ref{fig:circle}. In our experiments, we compare the performance of the classifiers to one another and characterize boundary conditions for each model where it exhibits rapid degradation in detection performance.

\section{Method}
\label{sect: method}
The development of accurate and robust deep learning models requires large amounts of diverse training data representing the environments where the models will be deployed; however, it is not always practical to collect real data in those environments. This could be due to the cost of the data collection, inclement weather, the timeframe in which the models are needed, or lack of access to the targets of interest. Therefore, the ability to augment existing real datasets with data from synthetic scenes constructed to simulate mission-relevant tasks and terrains is desirable. The synthetic scenes in this project were constructed using the Unity game engine, and C\# scripts were written to iterate through different combinations of camera positions relative to the target and different sun angles. The resulting data contained multiple targets and target poses and was automatically annotated.

\subsection{Virtual Environment}
\label{subsect:ve}
\begin{figure}[!htbp]
\centering
\includegraphics[width=0.95\textwidth]{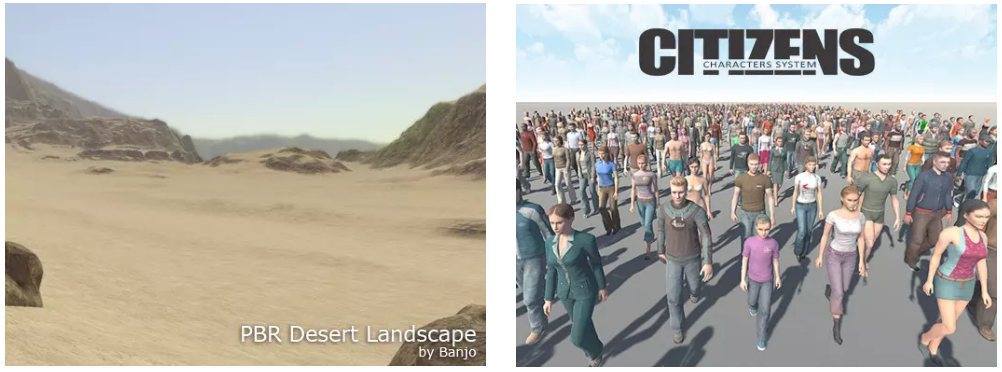}
\caption{PBR Desert Landscape asset from the game developer Banjo (facebook.com/BanjoGameArt) (Left), Citizen’s Pro 2019 asset package from the AGLOBEX-Mobile design team (aglobex.com) (Right)}
\label{fig:fig1}
\end{figure}

A game engine consists of a set of tools that allows a developer to construct virtual environments from individual, interacting game objects. The game engine enables the control of aspects of the game objects, such as their appearance and actions, and then renders the result. For this project, the Unity game engine was used to construct 3D scenes from game objects purchased from the Unity Asset Store. For the results shown below, a desert terrain asset\cite{PBR} and several human character assets\cite{CP2019} were used, as shown in Figure \ref{fig:fig1}.

\subsubsection{Simulation}
\begin{figure}[h]
\centering
\includegraphics[width=0.95\textwidth]{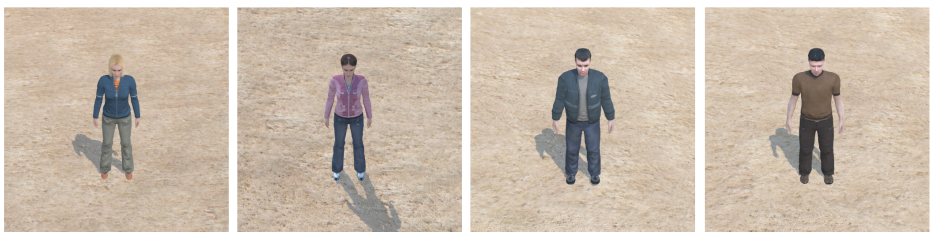}
\caption{Four different characters from the Citizen’s Pro 2019 asset package}
\label{fig:characters}
\end{figure}

Our process for generating a synthetic dataset starts with creating a Unity project and configuring the lighting and camera. A terrain asset and one or more target assets, such as those shown in Figure~\ref{fig:characters}, are then added to the terrain. The targets are the objects that we wish to annotate for use in training, testing, or validating deep learning models. Next, a unique tag is assigned to each target asset, and Rigidbody and Skinned Mesh Renderer components are attached. It is the tag that allows the targets to be independently segmented when they are later annotated.

\begin{figure}[h]
\centering
\includegraphics[width=0.98\textwidth]{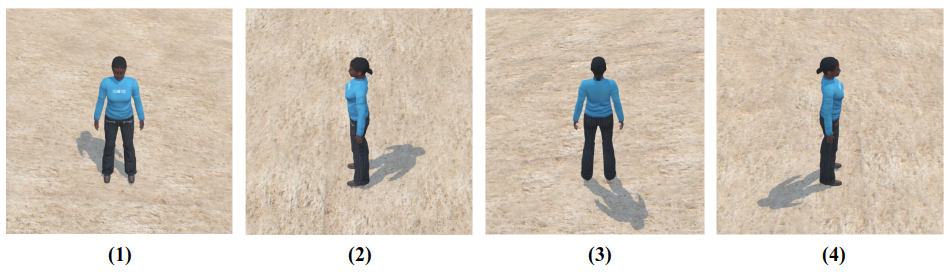}
\caption{The camera flies circling around the target and captures an image every 2\degree: (1) angle=0\degree, (2) angle=90\degree, (3) angle=180\degree, (4) angle=270\degree}
\label{fig:rotation}
\end{figure}

\begin{figure}[h]
\centering
\includegraphics[width=0.98\textwidth]{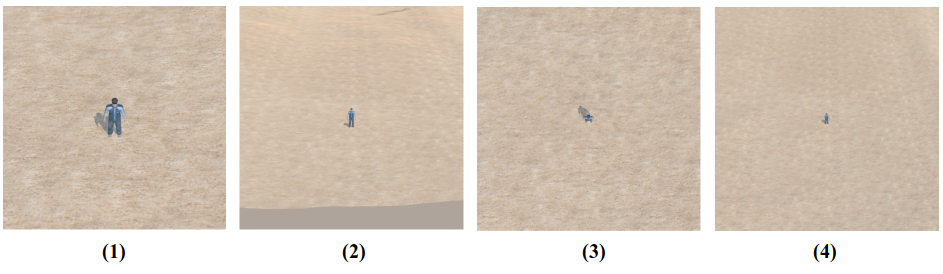}
\caption{The camera flies in circles of different radii around the target at different altitudes: (1) radius=20 m, altitude=25m, (2) radius=25m, altitude=40 m, (3) radius=35 m, altitude=25 m, (4) radius=50m, altitude=50 m}
\label{fig:range}
\end{figure}

A C\# script then controls the position and pose of a camera as it flies circling around the target. At each step, the camera is pointed at the center of the target (LookAt method). The script iterates the altitude of the camera, the radius of the circle, and the angle of the camera relative to the target to produce a range of poses, camera-to-target distances, and camera pitch angles, as shown in Figure \ref{fig:rotation} and \ref{fig:range}.

\begin{figure}[h]
\centering
\includegraphics[width=0.98\textwidth]{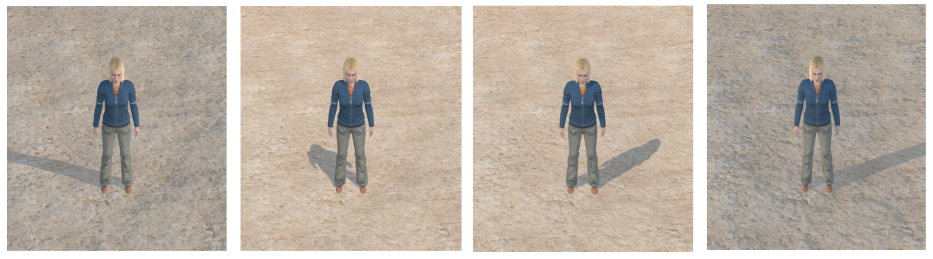}
\caption{Data was collected with different sun angles (different times of day)}
\label{fig:kelly_sunangle}
\end{figure}

Additionally, the sun angle was varied so that data could be generated at different times of day, as shown in Figure ~\ref{fig:kelly_sunangle}. This resulted in large sets of data that had variations in target position relative to the camera and time of day. For example, in one synthetic data generation trial, for a single virtual character, the altitude of the camera was varied from 5-50 meters in 5 meter increments, the radius of the circle was also varied from 5-30 meters in 5 meter increments, the angle of the camera relative to the character was varied from 0-358\degree \newline in 2\degree increments, and four different sun angles were simulated for a total of 43,200 images.

\begin{figure}[h]
\centering
\includegraphics[width=0.75\textwidth]{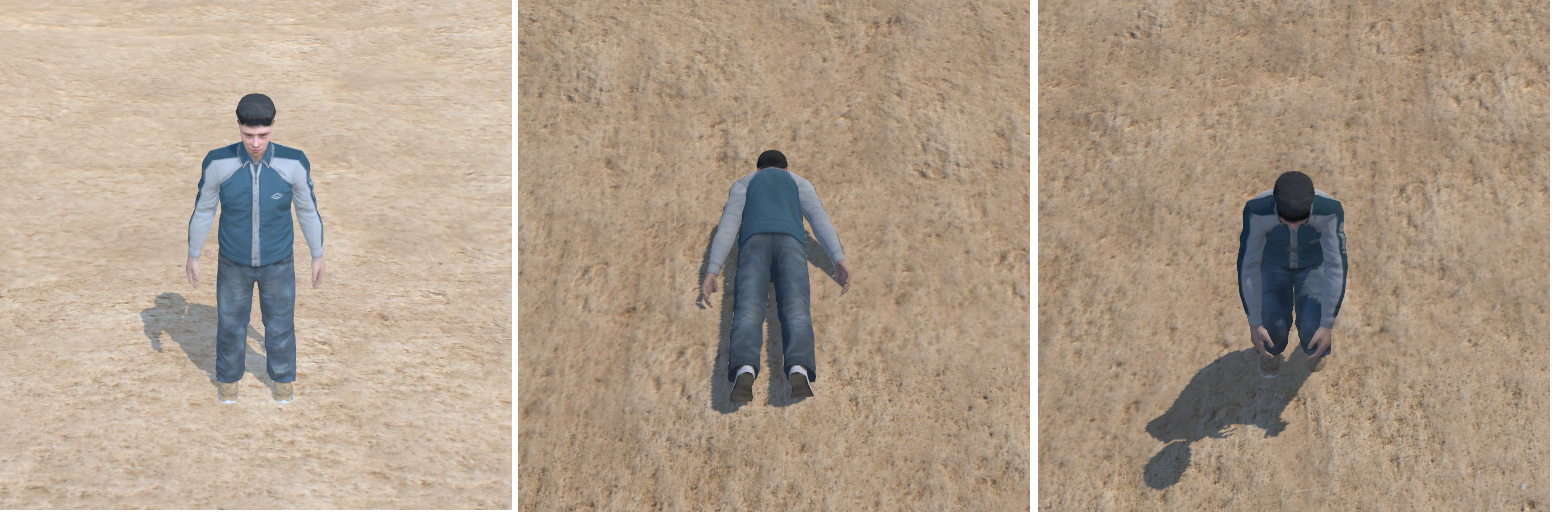}
\caption{Target poses: (Left) standing, (Center) prone, (Right) squatting}
\label{fig:pose}
\end{figure}

A synthetic data generation trial is a set of images that are produced without any user intervention after the initial setup. The initial setup involves selecting the terrain, the targets, and the target positions and poses. Our desert terrain dataset includes 8 different targets, each in 3 different poses. The 3 poses are standing, prone, and squatting,  as shown in Figure \ref{fig:pose}. Therefore, the total number of images across the 24 trials is about 1 million.

\subsection{Automatic Annotation}
\begin{figure}[h]
\centering
\includegraphics[width=0.98\textwidth]{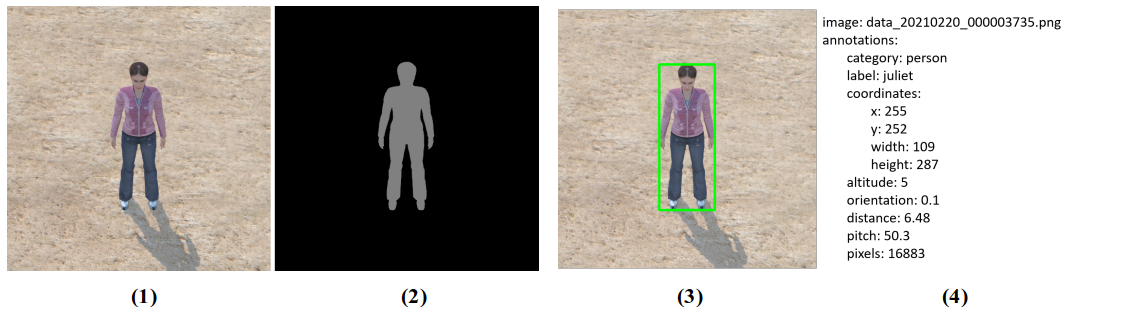}
\caption{(1) Synthetic image, (2) Image segmentation mask, (3) Training data image with bounding box, (4) Annotations associated with the image}
\label{fig:mask_bb}
\end{figure}

To synthesize the images from the virtual scene, we used a repository called Image Synthesis for Machine Learning\cite{U3DC}. The repository contains C\# code for generating annotated training sets in Unity. Specifically, the code produces an image segmentation mask where each object is assigned a unique color and the synthetic image, as shown in Figure \ref{fig:mask_bb}.


To generate the annotations, we have written a Python script to parse each mask file, identify each target, and measure the center coordinates, width, and height of the smallest rectangle that encompasses the target, as shown in Figure \ref{fig:mask_bb}. Additionally, the object label, label category, camera altitude (m), the orientation of the target relative to the camera (degrees), distance of the target from the camera (m), the pitch angle of the camera (degrees), and the number of pixels included in the image segmentation mask are recorded in a single JSON file for each trial.


\subsection{Experimental Setup}
\subsubsection{Experimental Datasets}
As described in section ~\ref{subsect:ve}, we populate synthetic datasets in various camera positions, viewing angles, illumination conditions, and target object appearances and poses. The image frame consists of a 512 × 512 pixel array with a background terrain containing  desert and mountains, and the target object is placed in the center of each image frame. For our experiment, we use a person class to validate machine learning models. We validated the detection performance of machine learning models as a function of the various image parameters. In our experiment, we use average precision (AP) of an IOU (Intersection of Union) threshold of 50\% as the indicator of detection performance.
 
\subsubsection{Detection Models}
For detectors, we use three well-known machine learning models: Tiny-YOLO, YOLOv3, and RetinaNet. These models are representative one-stage detectors that can achieve high-speed inference. Using multiple models, we characterize the performance of detection models, which have different levels of architectural and model complexity.

\section{Experimental Results}
\label{sect:results}
We visualize detection performance for each model for a given distance and view angle between the camera and the target object, and compare the performance of the models to one another in different target poses. The analysis is to demonstrate if a trained detector has a bias toward specific poses. Additionally, we present detection performances in four different illumination conditions that are determined by the sun angles. In this work, we refer to four illumination conditions as follows: early morning, around noon, mid-afternoon, and late afternoon. We also present a histogram that describes how detection accuracy changes as the camera flies in circles around the target object as shown in Figure ~\ref{fig:rotation}.

\subsection{Standing Position}
\subsubsection{Overall detection performance}
\begin{figure}[!htbp]
\centering
\includegraphics[width=0.92\textwidth]{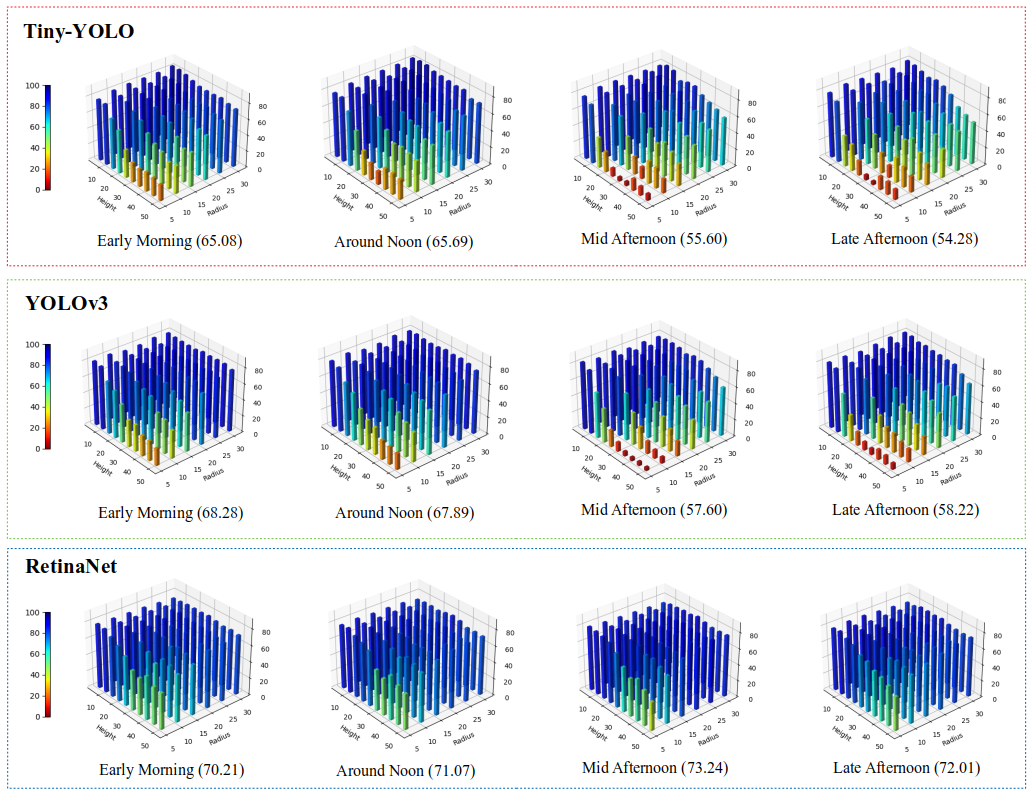}
\caption{Detection performance of the three models in the standing position with different illumination conditions. Each bar in the plots indicates an AP value at the corresponding camera height and radius of the circle. The numbers in the parentheses are mAP (mean Average Precision) values.}
\label{fig:stand}
\end{figure}

Figure ~\ref{fig:stand} illustrates detection performance for the standing position in different imaging conditions. In the figure, detection accuracy is binned by the camera height and the radius of the circular path of the camera as it moves around the object within the image. We also provide the mean of AP values in each illumination condition for performance comparisons.

\begin{figure}[!htbp]
\centering
\includegraphics[width=0.5\textwidth]{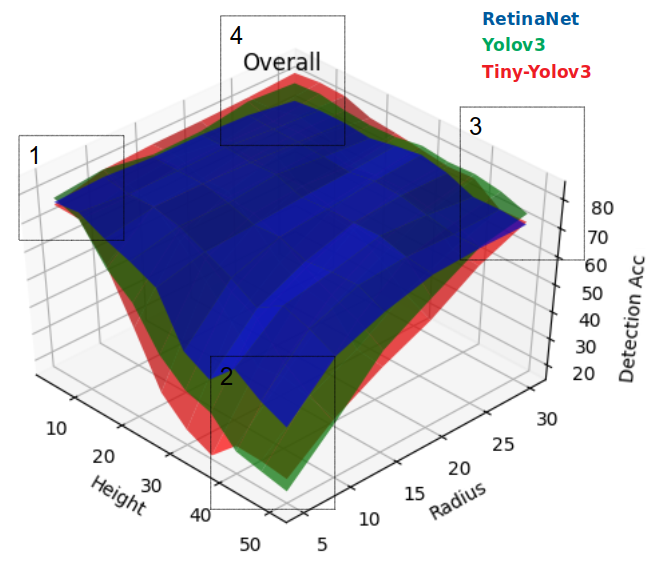}
\caption{Overall detection performance in the standing position. 1) large target object, 2) nadir-view image, 3) small target object, 4) eye-level-view image}
\label{fig:stand-all}
\end{figure}

Figure ~\ref{fig:stand-all} shows the overall detection accuracy from the three detectors in the standing position. The detection results of all three models are unsatisfactory for images that are captured with a high viewing angle (approaches the nadir view). This may be partly the result of the training dataset not containing nadir or near-nadir views, or bird-eye views, of image scenes captured by the camera at high altitudes and on the circles of small radii. In the same dataset, the detection results for images captured from short distances have relatively high AP values. This is because the target object takes up a large percentage of the image frame. Additionally, the detection accuracy increases for a given altitude as the camera is moved farther away. This is likely because the pre-trained detection model uses ground imagery, so the detection accuracy increases as the images appear more similar to the image characteristics of ground imagery. Figure ~\ref{fig:stand-all} shows that the overall performance of the three models that were compared and indicates that RetinaNet is more robust than that of YOLO3 and Tiny-Yolo in general due to advanced model architecture requiring more computational resources.

The overall detection results can be classified into four large regions: 1) large target images, 2) nadir-view images, 3) small target images, and 4) eye-level-view images. In the first region, the target objects are large where the image frames are captured at a low altitude and small radius. In another region, target objects are small where the image frames are captured at a high altitude and large radius. Nadir-view images are those captured when the camera is at a high altitude and the circle radius is small. Eye-level-view images are those captured when the camera is at a low altitude and the circle radius is large and are characterized by small pitch angles. Image frames from the eye-level view are well covered by large benchmark datasets, resulting in high overall detection performance.

\subsubsection{Angular dependency plot}
\begin{figure}[!htbp]
\centering
\includegraphics[width=0.95\textwidth]{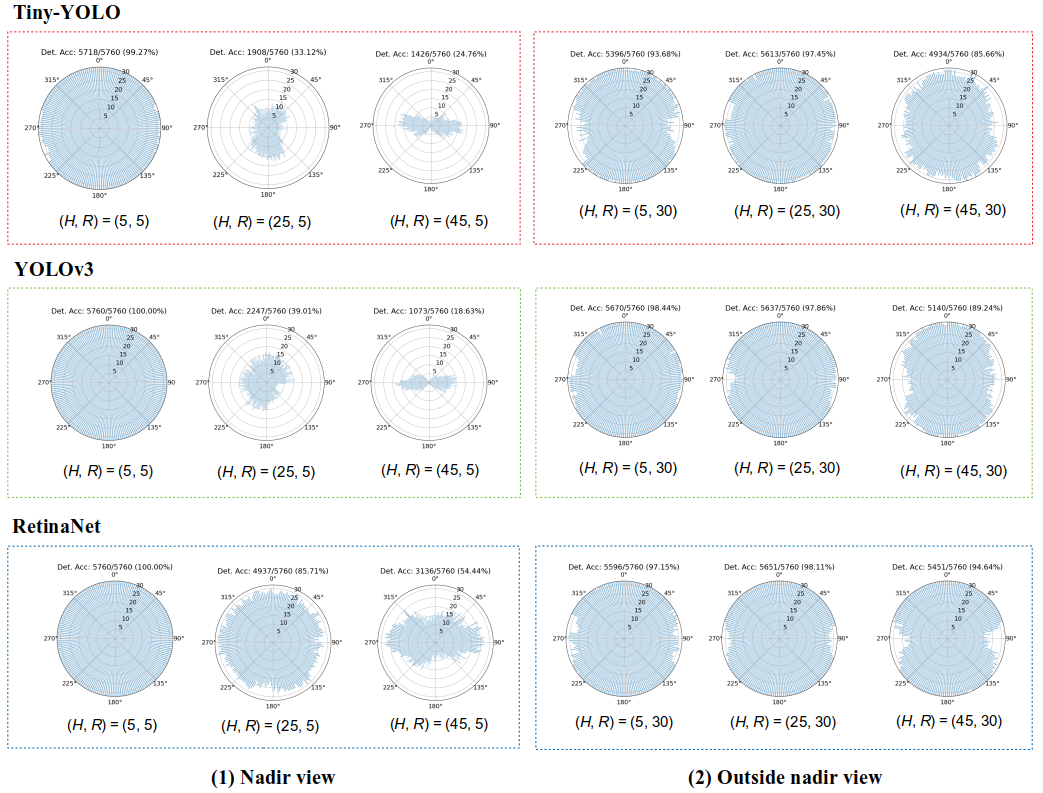}
\caption{Angular dependency plots for the standing position: H and R indicate height and radius, respectively. 1) nadir view, 2) outside nadir view}
\label{fig:stand-angle}
\end{figure}

Figure ~\ref{fig:stand-angle} is a histogram illustrating the number of positive detections at each camera view angle. The only variable that changes in each histogram is the camera angle relative to the object within the image as the camera moves along the circle. There is a bias in the detection results towards the person in the front and back view, which may indicate that a similar image dataset is included in the dataset used for training the model.

\subsection{Squatting Position}
\subsubsection{Overall detection performance}
\begin{figure}[!htbp]
\centering
\includegraphics[width=0.90\textwidth]{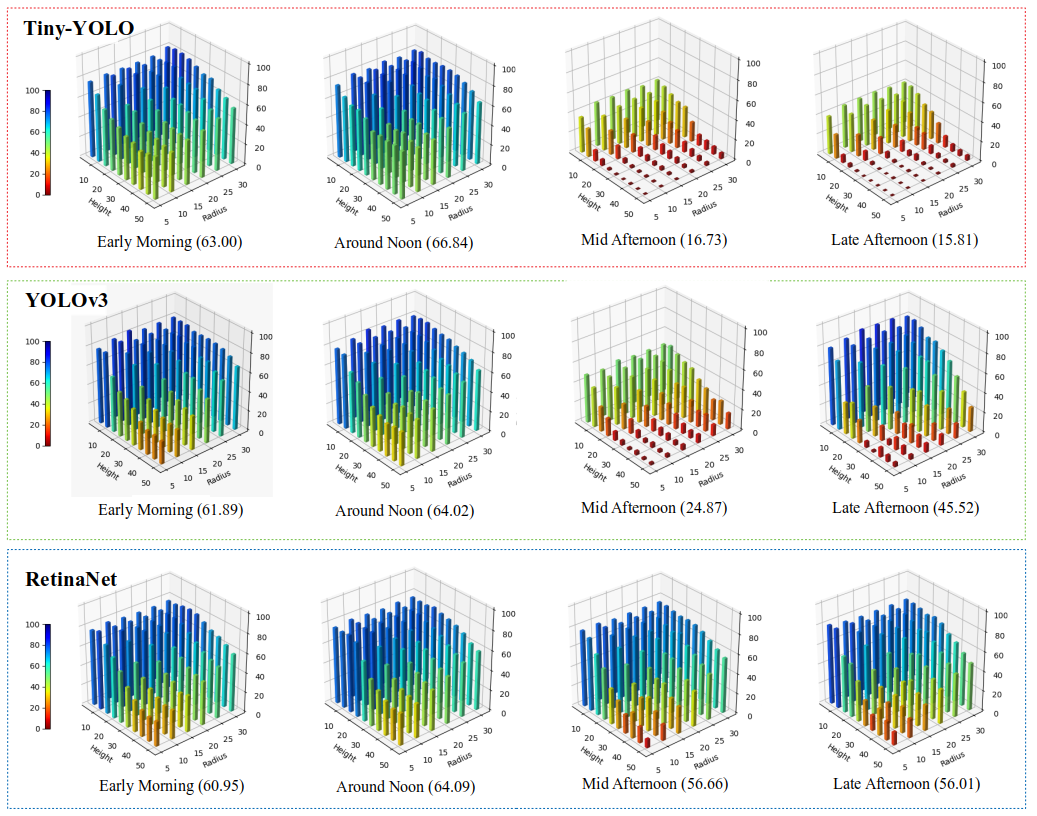}
\caption{Detection performances of three models in the squatting position with different illumination conditions. Each bar in the plots indicates an AP value at the corresponding camera height and radius of the circle. The numbers in the parentheses are mAP (mean Average Precision) values.}
\label{fig:squat}
\end{figure}

Figure ~\ref{fig:squat} shows the detection performance in the squatting position with four different illumination conditions.

\begin{figure}[!htbp]
\centering
\includegraphics[width=0.5\textwidth]{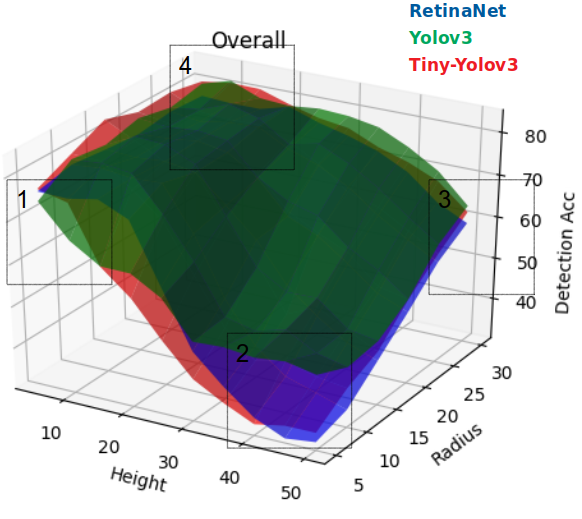}
\caption{Overall detection performance in the squatting position. 1) large target object, 2) nadir-view image, 3) small target object, 4) eye-level-view image}
\label{fig:squat-all}
\end{figure}

As presented in Figure ~\ref{fig:squat-all}, overall detection performance in the squatting position has similar results to those observed for the standing position as a function of height and radius parameters. However, the overall detection accuracy is lower because the size of target object is smaller in the squatting position, which degrades the detection performance, especially in long-distance and steep viewing angle conditions.

\subsubsection{Angular dependency plot}
\begin{figure}[!htbp]
\centering
\includegraphics[width=0.90\textwidth]{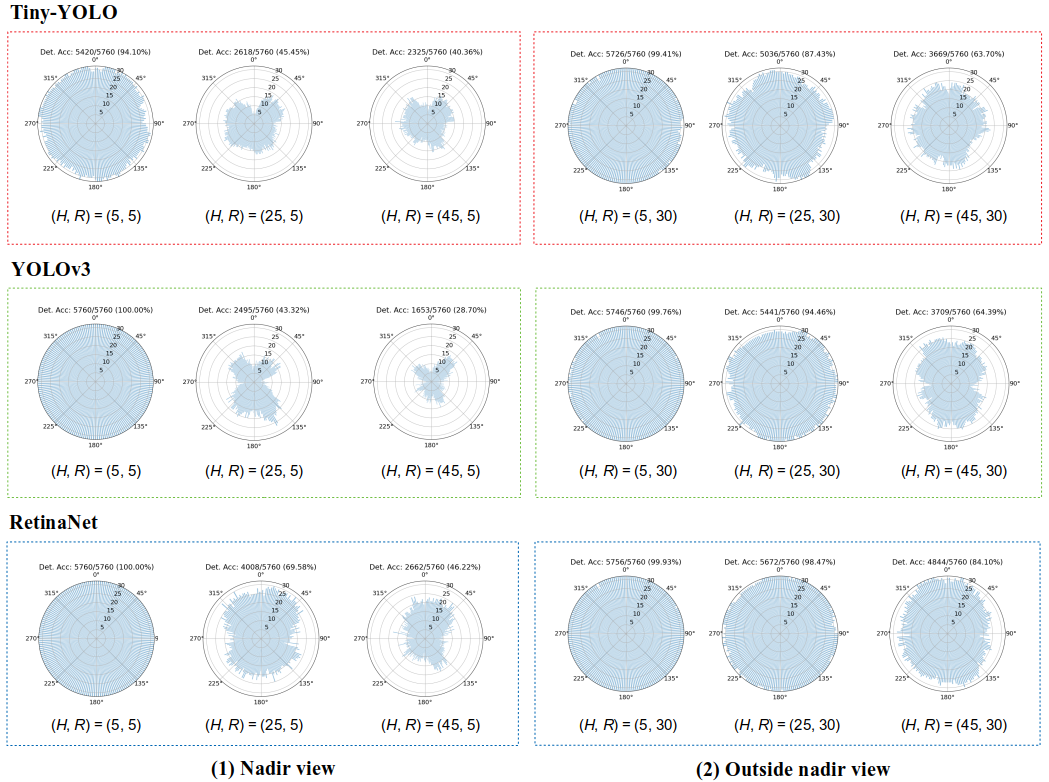}
\caption{Angular dependency plots for the squatting position: H and R indicate height and radius, respectively. 1) nadir view, 2) outside nadir view}
\label{fig:squat-angle}
\end{figure}

Figure ~\ref{fig:squat-angle} presents the number of positive detection at each camera view angle in the squatting position. 

\subsection{Prone Position}
\subsubsection{Overall detection performance}
\begin{figure}[!htbp]
\centering
\includegraphics[width=0.90\textwidth]{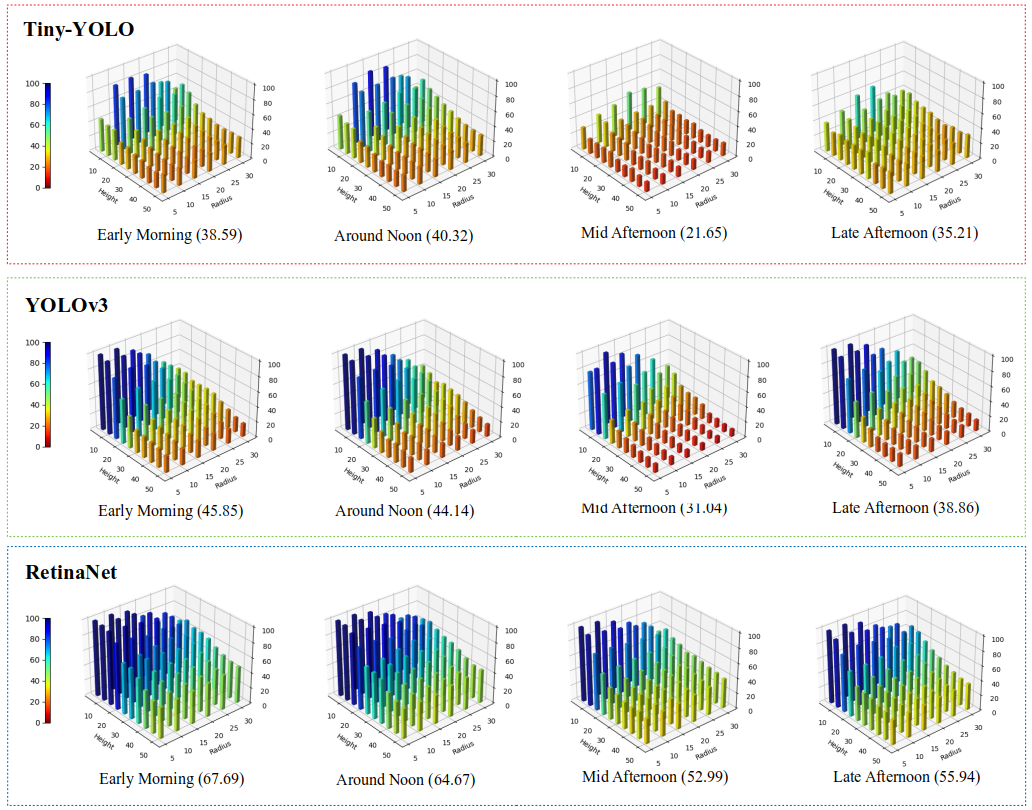}
\caption{Detection performances of three models in the prone position with different illumination conditions. Each bar in the plots indicates an AP value at the corresponding camera height and radius of the circle. The numbers in the parentheses are mAP (mean Average Precision) values.}
\label{fig:lye}
\end{figure}

Figure ~\ref{fig:lye} presents the detection performance of three models in the prone position with different illumination conditions.

\begin{figure}[!htbp]
\centering
\includegraphics[width=0.5\textwidth]{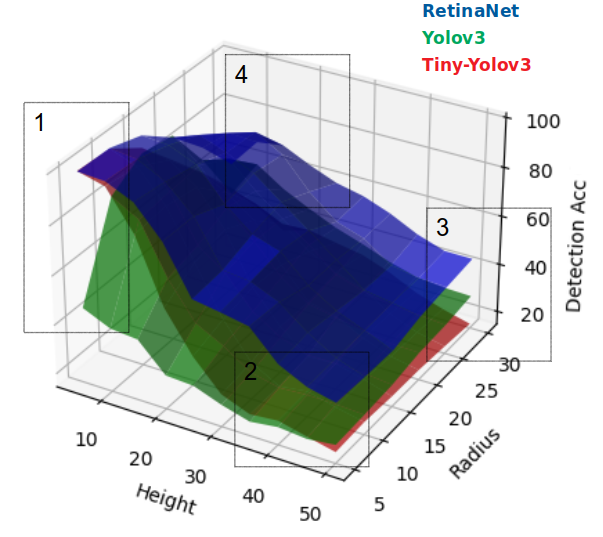}
\caption{Overall detection performance in the prone position. 1) large target object, 2) nadir-view image, 3) small target object, 4) eye-level-view image}
\label{fig:lye-all}
\end{figure}

Figure ~\ref{fig:lye-all} illustrates the overall detection accuracy of three models in the prone position. As illustrated in Figure ~\ref{fig:lye-all}, the prone position detection performance over the non-nadir-view region shows a significant reduction in accuracy compared to that of the standing and squatting positions. This is primarily because the prone position is a relatively unique pose unfamaliar to the neural network models pretrained on large-scale benchmarks. 

\subsubsection{Angular dependency plot}
\begin{figure}[!htbp]
\centering
\includegraphics[width=0.95\textwidth]{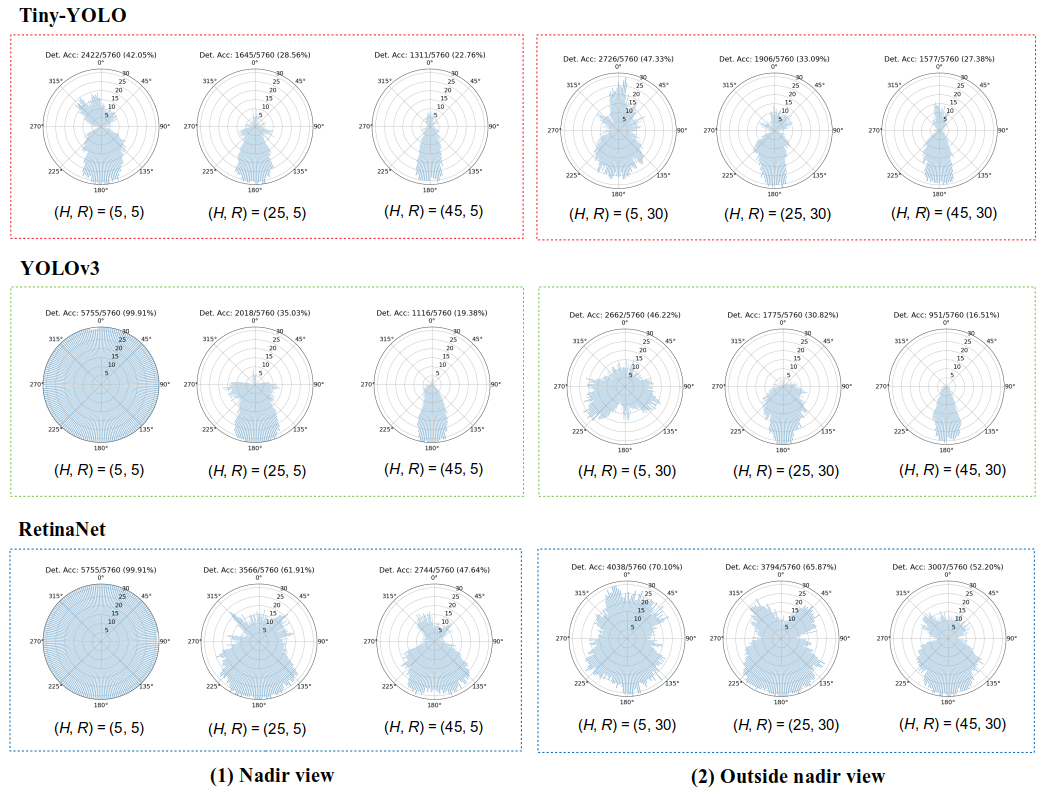}
\caption{Angular dependency plots for the prone position: H and R indicate height and radius, respectively. 1) nadir view, 2) outside nadir view}
\label{fig:lye-angle}
\end{figure}

Figure ~\ref{fig:lye-angle} presents detection results for angular dependency in the prone position. We observe that there are biased detection results towards the 180 \degree position both in nadir view and outside of nadir view image frames. This is because the target object looks like the person in the back-side view, which is contained in the training dataset.

Our analysis has shown that the detection performance of all three models are lower for image frames collected directly above the target object. The results indicate that the models should be retrained if the goal is to use them with aerial imagery. We also observe that lower illumination conditions degrade the overall detection performance.

\section{Conclusion}
\label{sect:misc}
In this study, we generate synthetic data using a game engine to characterize detection accuracy of machine learning models  in various conditions of UAV-based imaging systems. By applying neural network models with different model complexity to the synthetic data, we quantitatively show how detection accuracy varies as imaging conditions change. Additionally, we characterize boundary conditions for the neural network models beyond which the models exhibit rapid degradation in detection accuracy. The proposed work provides valuable information regarding the accuracy and usability of UAV-based classifiers onboard UAVs at the edge by characterizing practical limits under which the classifiers can be reliably applied. Also, our analysis presented in this study will allow the user to select the optimal classifier for a given set of imaging parameters.

\section{Acknowledgment}
\label{sect:ack}
This research was sponsored by the Defense Threat Reduction Agency (DTRA). The views and conclusions contained in this document are those of the authors and should not be interpreted as representing the official policies, either expressed or implied, of the Army Research Laboratory or DTRA.

\bibliographystyle{spiebib} 
\bibliography{main} 

\end{spacing}
\end{document}